\pdfoutput=1

\documentclass[11pt]{article}

\usepackage[]{acl}

\usepackage{times}
\usepackage{latexsym}

\usepackage[T1]{fontenc}

\usepackage[utf8]{inputenc}

\usepackage{microtype}

\usepackage[linesnumbered, ruled]{algorithm2e}
\usepackage[T1]{fontenc}
\usepackage{dsfont}

\usepackage{mdframed}
\usepackage{booktabs}
\usepackage{graphicx}

\usepackage[normalem]{ulem}
\usepackage{enumitem}
\usepackage{xcolor}
\usepackage{soul}
\colorlet{soulred}{red!30}
\sethlcolor{soulred}%

\usepackage{amsmath,amsthm,amsfonts,amssymb,bm}
\usepackage{framed}
\usepackage{varioref}
\usepackage{float}
\usepackage{multirow}
\usepackage{dashrule}
\usepackage{url}
\usepackage{pifont}
\usepackage{lipsum}
\usepackage{algorithmic}
\usepackage{amsmath}

\usepackage{multicol}

\title{Knowledge-grounded Dialog State Tracking}

\author{Dian Yu \thanks{work done while at University of California, Davis}, Mingqiu Wang, Yuan Cao, Izhak Shafran, Laurent El Shafey, Hagen Soltau\\
Google Research\\
\texttt{\{dianyu, mingqiuwang, yuancao, izhak, shafey, soltau\}@google.com}
}

\begin{document}
\maketitle
\begin{abstract}
Knowledge (including structured knowledge such as schema and ontology, and unstructured knowledge such as web corpus) is a critical part of dialog understanding, especially for unseen tasks and domains. Traditionally, such domain-specific knowledge is encoded implicitly into model parameters for the execution of downstream tasks, which makes training inefficient. In addition, such models are not easily transferable to new tasks with different schemas. In this work, we propose to perform dialog state tracking grounded on knowledge encoded externally. We query relevant knowledge of various forms based on the dialog context where such information can ground the prediction of dialog states. We demonstrate superior performance of our proposed method over strong baselines, especially in the few-shot learning setting.

\end{abstract}

\section{Introduction}
Pre-trained language models (LMs, \citealp{gpt2, t5}) are the backbone of contemporary task-oriented dialog (TOD) models \cite{peng-etal-2020-soloist, yang-etal-2021-ubar}. However, the models are pre-trained on large generic corpora so that they do not contain task-specific knowledge. 
Previous work primarily suggests further pre-training or fine-tuning the LMs on in-domain data for adaptation \cite{wu-etal-2020-tod, simpletod}, but it cannot consider information above the surface level. This makes it challenging for downstream tasks especially in the few-shot learning setting because mapping representation to the output space and encoding knowledge into the model parameters are entangled, while the latter may require more training data.
Some more recent research proposes to incorporate external knowledge for response generation tasks \cite{dinan-etal-2019-wizard, shuster-etal-2022-language, chen-etal-2022-ketod, komeili-etal-2022-internet}, but it is not clear how to utilize such information for language understanding. 


\begin{figure}[t]
\centering
\includegraphics[width=\linewidth]{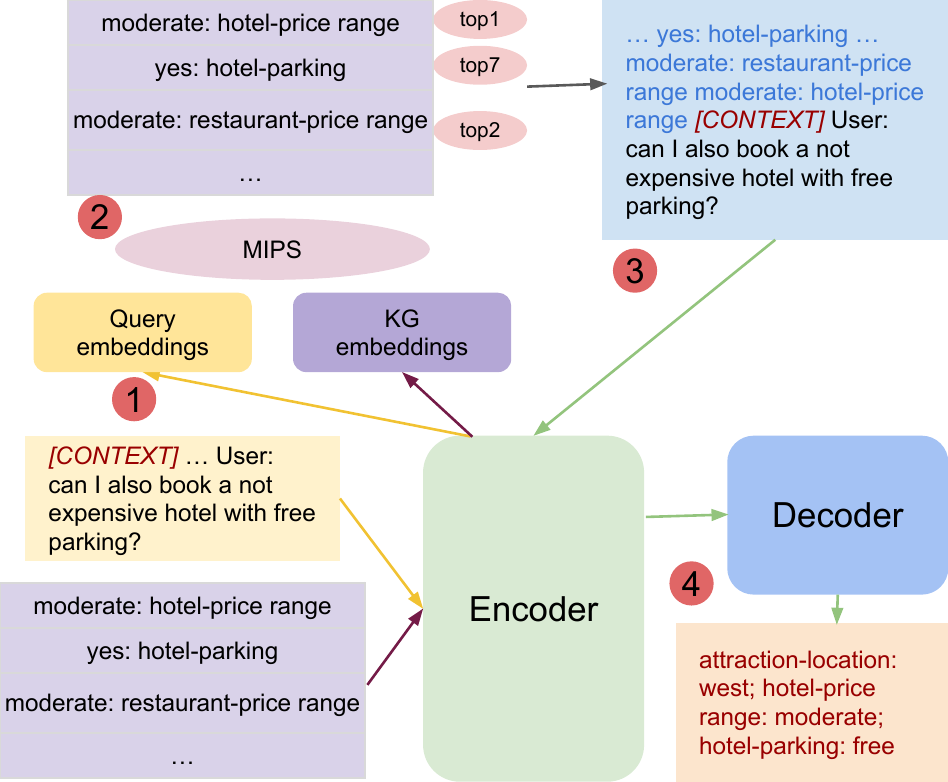}
\vspace{-1em}
\caption{Model architecture for our proposed knowledge-grounded DST. The encoder first encodes the query and knowledge into representations, and we find the top-k most relevant knowledge elements to the context in step 2. We flatten the retrieved elements in step 3 and append to the query context as the input to the encoder-decoder model. The retrieved elements serve as a prior for DST.}
\label{fig:model}
\vspace{-1em}
\end{figure}

In TOD settings, because the API call structure is restricted to certain intents, slots, and values, the schema is often provided. For example, in a flight booking system, queries like departure location and airlines are pre-defined. Users, even though not bounded directly by what they can say to agents, have a limited and predictable vocabulary set to some extent. If the schema information is utilized, a model does not need to learn that “San Francisco” represents a departure place, rather than a general city name from the LM. This is particularly important for new information, such as movie titles or locations that do not appear in the LM training corpus.
Similar to human annotators, grounding a dialog model on such knowledge makes it easier and more accurate in understanding conversations.

In this paper, we investigate knowledge-grounded understanding for dialog state tracking (DST). In addition to using structured knowledge such as the ontology of slot type-value pairs, we also consider unstructured knowledge from the raw training data. We train a TOD model to query relevant knowledge for each turn in the context, and leverage the retrieved knowledge to predict dialog state. 
We evaluate our method on MultiWOZ \cite{budzianowski-etal-2018-multiwoz} for both the full-data and few-shot settings, and show superior performance compared to previous methods.  

\section{Related Work}

\subsection{Knowledge grounding}

To relax the requirement of encoding knowledge of the whole world into model parameters,
one direction 
is to disentangle knowledge representation from LMs. Most of these methods are applied to knowledge-intensive text generation tasks such as open-domain question answering \cite{lee-etal-2019-latent, 
karpukhin-etal-2020-dense, guu-etal-2020-retrieval, 
lewis-etal-2020-retrieval, boregeaud-etal-2021-improving}, and response generation with factual information \cite{dinan-etal-2019-wizard, 
komeili-etal-2022-internet, lamda, 
kim-etal-2020-beyond, thulke-etal-2021-efficient, chen-etal-2022-ketod}. 
Similarly, some work also considers retrieving information to serve as a reference to refine the model generation process \cite{weston-etal-2018-retrieve, gonzalez-etal-2019-retrievalbased, khandelwal-etal-2021-nearest, zhang-etal-2021-jointRetrieval}.
Different from these approaches, our method focuses on learning and utilizing available domain-relevant knowledge for language understanding tasks. Moreover, we propose to leverage knowledge of various formats.

\subsection{Knowledge guided dialog understanding}

Encoding domain schema into model parameters \cite{simpletod, madotto-etal-2020-learning} may not be efficient for unseen domains and tasks where the ontology can be different. 
One line of research \cite{ren-etal-2018-towards, wu-etal-2019-transferable,  zhou-small-2019-multi, rastogi-etal-2020-towards, du-etal-2021-qa, lee-etal-2021-dialogue}
leverages question-answering techniques to predict values for each slot, or prepend all slot-value information to the context \cite{zhao-etal-2022-description}.
However, this method is not scalable when the number of slot-value pairs is large, especially in multi-domain TOD systems. In addition, probably due to blurry attention over long context \cite{fan-etal-2021-augmenting}, \citet{lee-etal-2021-dialogue} find that adding potential slot values does not improve the model performance. In contrast, retrieving only relevant schema effectively solves the scalability problem by specifying the knowledge with a fixed length.

Alternatively, instead of structured schema knowledge, recent research proposes to use hand-crafted demonstrations as prompts \cite{gupta-etal-2022-show} or find similar examples to guide understanding tasks \cite{yu-etal-2021-shot, pasupat-etal-2021-controllable, yao-etal-2021-tlm} such as conversational semantic parsing. However, one turn can contain multiple dialog states so that retrieved examples from previous methods may not be sufficient to provide required evidence. Furthermore, our method can be applied to unify different forms of knowledge including structured and unstructured ones.


\section{Methodology}
Our proposed method is illustrated in Figure \ref{fig:model}. Given the context $\mathbf{x}$, we first retrieve $\mathbf{k}$ relevant knowledge entries $\mathbf{e}$ by the similarity between $\operatorname{Enc}(\mathbf{x})$ and $\operatorname{Enc}(\mathbf{e})$ using an encoder $\operatorname{Enc}$. Then we integrate the retrieved entries $e_1, e_2, ..., e_k$ with the original context to form $\mathbf{x'}$, where $\mathbf{x'}$ is used as the input for the target DST task.

\paragraph{Knowledge retrieval}
Different from previous work (such as question answering) where there is only one ground-truth knowledge for each query, 
multiple entries of the form slot-value pairs may exist in the ontology base that match the conversation context.
Importantly, unlike passage retrieval where the query (e.g., a sentence) and the target (e.g., another sentence or passage) are similar to the pre-training corpus, structured knowledge such as schema pairs may have different representation distribution. Thus, an off-the-shelf encoder may retrieve noisy elements and degrade final performance, especially when training with the target task optimized on DST generation. Moreover, non-parametric retrieval methods such as TF-IDF and BM25 \cite{robertson-zaragoza-2009-bm25} rely on lexical overlapping, which could be detrimental when entries in schemas contain high word overlapping (e.g., same value for different slots).

We therefore train our knowledge retriever to promote similar representations between a query and its ground truth knowledge. We started with optimizing the marginal likelihood over all positive knowledge entries, but found that it resulted in peaky distribution centered around specific elements in our preliminary studies. Instead,
we minimize binary cross-entropy with contrastive loss:

\begin{small}
\begin{equation}
\small{
\begin{split}
    \mathcal{L} &=
    -\sum_{i=1}^{i=n}y_i\cdot 
    (log(sim(\operatorname{Enc}(\mathbf{x}),\operatorname{Enc}(e_i))) \\ 
    &+(1 - y_i)\cdot log(1 - sim(\operatorname{Enc}(\mathbf{x}),\operatorname{Enc}(e_i))))
\end{split}
}
\end{equation}
\end{small}

\noindent where $y_i$ is 1 if $e_i$ appears in the target dialog state and otherwise 0. 
In our model, we use the same encoder $\operatorname{Enc}$ for both the context and the knowledge, and $\operatorname{Enc}$ is also used for the target DST task. 
$sim$ defines the retrieval score,
computed as the dot product between representations of the first token from the last layer\footnote{Other methods such as ColBERT \cite{khattab-etal-2020-colbert} are also applicable. In our preliminary experiments, we found that dot product is an effective measure, corroborating findings from \citet{ni-etal-2021-large}}.

\paragraph{Knowledge integration}
Once most relevant knowledge elements are retrieved by the model, this extra information can serve as a strong inductive bias to the downstream, knowledge-sensitive tasks. One common approach for knowledge integration is fusion-in-decoder \cite{izacard-grave-2021-distilling}. Although efficient, it has been shown that retrieved information is likely to be ignored by a pre-trained model \cite{shuster-etal-2022-language}. Hence, we concatenate retrieved knowledge with the context $\mathbf{x'} = e_k, e_{k-1}, ..., e_1, \mathbf{x}$, where the entries are ordered from the least similar ($e_k$) to the most similar ($e_1$). The similarity can also be considered as the confidence an element $e_i$ is relevant to the current context. We take the $\mathbf{x'}$ as the context to the DST task. Therefore, our method is unified for knowledge of any format, and a bounded number of elements 
can solve the problem of memory constraint in previous research \cite{zhao-etal-2022-description}.

\section{Experiments and Results}
\subsection{Experiments and baselines}
We conduct experiments on MultiWOZ 2.4 \cite{budzianowski-etal-2018-multiwoz, ye-etal-2021-multiwoz} for DST in both full-shot and few-shot (1\%) learning settings. For all experiments, we use T5-base and T5-XXL encoder-decoder models \cite{t5} as the initial checkpoints. 
We use the publicly available T5 checkpoints \footnote{\url{2https://github.com/google-research/ text-to-text-transfer-transformer/blob/ main/released_checkpoints.md}} for our experiments. T5-base has 250 million parameters, and T5-XXL has 11B parameters. We train all models on 64 (for T5-base) and 128 (for T5-XXL) TPU v3 chips \cite{jouppi-etal-2017-in}. For fine-tuning, we set a learning rate of 1e-4 and a batch size of 32. We set the input and output sequence length to 1024 and 512 tokens. We train all models for 200k steps and report the performance on the test set from checkpoints achieving the best results on the development set. When multitask training on both retrieval and DST, we set 0.1 weight to the retrieval loss and 1 weight to the DST loss since it is relatively faster to converge for retrieval. We also experimented with 0.01, 0.05, 0.5, 1 for the retrieval loss weight, and found that 0.1 performs the best.

We compare with two baselines, seq2seq and D3ST. seq2seq takes the context as input, and predicts a sequence of linearized dialog state for each turn. Similarly, D3ST \cite{zhao-etal-2022-description} adds descriptions of each slot with potentially values as the prompt and predicts dialog states as multiple choice. Both baselines use the same T5 initial checkpoints. 
We report averaged joint goal accuracy (JGA) across three random seeds.

For our proposed method, we consider slot type (type), slot type and value (type+value), and training data (training) as knowledge sources. 
Specifically, for type, we consider all slot types (35 in total such as ``hotel-parking'') as the knowledge base and retrieve corresponding top ten elements. For type+value, we consider each combination of types and their values in the form of ``type: value'' (1858 in total such as ``hotel-parking: don't care'') as the knowledge elements. Because there are more elements, we consider top 30 in our experiments for retrieval to achieve higher recall (with analysis later in Section \ref{analysis}). For training data, because of memory concerns,  we randomly sample 500 training examples as the knowledge base and we consider the ground truth training example as the one with the highest F1 overlapping in the dialog slot types. We only consider top-1 due to the length constraint. For each knowledge source, we train retrieval together with the DST generator using the same model parameters. 

\subsection{Results}
\begin{table}[]
\small
\begin{center}
\resizebox{\columnwidth}{!}{
\begin{tabular}{cclccc}
\toprule
\multicolumn{1}{l}{}                       & \multicolumn{1}{l}{}                               & model      & JGA            & p          & r     \\ 
\midrule
\multicolumn{1}{c|}{\multirow{10}{*}{xxl}} & \multicolumn{1}{c|}{\multirow{5}{*}{1\%}}          & baseline   & 50.24          & -          & -     \\
\multicolumn{1}{c|}{}                      & \multicolumn{1}{c|}{}                              & D3ST       & 54.37          & -          & -     \\
\multicolumn{1}{c|}{}                      & \multicolumn{1}{c|}{}                              & type       & 53.59          & 38.08      & 99.30 \\
\multicolumn{1}{c|}{}                      & \multicolumn{1}{c|}{}                              & type+value & \textbf{55.32} & 12.94      & 48.78 \\
\multicolumn{1}{c|}{}                      & \multicolumn{1}{c|}{}                              & training   & 51.38          & 57.81      & 45.19 \\ 
\cmidrule{2-6} 
\multicolumn{1}{c|}{}                      & \multicolumn{1}{c|}{\multirow{5}{*}{full (100\%)}} & baseline   & 73.18          & -          & -     \\
\multicolumn{1}{c|}{}                      & \multicolumn{1}{c|}{}                              & D3ST       & \textbf{75.90} & -          & -     \\
\multicolumn{1}{c|}{}                      & \multicolumn{1}{c|}{}                              & type       & 73.72          & 38.30      & 99.80 \\
\multicolumn{1}{c|}{}                      & \multicolumn{1}{c|}{}                              & type+value & 75.47          & 12.49      & 68.56 \\
\multicolumn{1}{c|}{}                      & \multicolumn{1}{c|}{}                              & training   & 73.96          & 81.30      & 63.02 \\ 
\bottomrule
\multicolumn{1}{l|}{\multirow{8}{*}{base}} & \multicolumn{1}{c|}{\multirow{4}{*}{1\%}}          & baseline   & 30.48          & -          & -     \\
\multicolumn{1}{l|}{}                      & \multicolumn{1}{c|}{}                              & D3ST       & 16.37          & -          & -     \\
\multicolumn{1}{l|}{}                      & \multicolumn{1}{c|}{}                              & type       & 32.76          & 26.41      & 73.43 \\
\multicolumn{1}{l|}{}                      & \multicolumn{1}{c|}{}                              & type+value & \textbf{34.89} & 6.17       & 21.95 \\ 
\cmidrule{2-6} 
\multicolumn{1}{l|}{}                      & \multicolumn{1}{c|}{\multirow{4}{*}{full (100\%)}} & baseline   & 67.10          & -          & -     \\
\multicolumn{1}{l|}{}                      & \multicolumn{1}{c|}{}                              & D3ST       & \textbf{72.10} & \textbf{-} & -     \\
\multicolumn{1}{l|}{}                      & \multicolumn{1}{c|}{}                              & type       & 70.51          & 37.64      & 98.80 \\
\multicolumn{1}{l|}{}                      & \multicolumn{1}{c|}{}                              & type+value & 71.54          & 8.21       & 29.20\\
\bottomrule
\end{tabular}
}
\end{center}
\caption{\label{results} Dialog state tracking results on MultiWOZ. We report averaged joint goal accuracy and retrieval metrics (precision and recall). With both T5-base and T5-xxl, grounding on retrieved slot types and values achieves better results by a large margin on the 1\% few-shot learning setting, while performing on par with the full data setting. }
\vspace{-1em}
\end{table}

Table \ref{results} shows DST results produced by different methods. Compared to the seq2seq baseline and D3ST, grounding on relevant knowledge by retrieval achieves better JGA by a large margin especially in the few-shot learning setting ($> 4\%$ absolute value). In the full data setting, our method performs on par with D3ST mostly due to that with more training data, the model can encode knowledge into parameters rather than relying on a separate, disentangled knowledge base. However, our method is not limited by the sequence length when we can specifically choose the number of retrieved elements regardless of the ontology size. 

When comparing among different knowledge formats, type+value performs better than retrieving type only despite that retrieving is a harder task. As shown by recall\footnote{Precision is determined by the number of retrieved elements we set, whereas recall measures the percentage of ground-truth knowledge elements being correctly retrieved. Therefore, recall is more informative.}, with a large pre-trained model (XXL), recall for retrieving type only can achieve perfect scores ($>99\%$), but recall for type+value can only be 48\% in the few-shot and close to $70\%$ in the full-data setting. This indicates that 
the model can denoise distracting elements and make use 
of relevant knowledge as a positive inductive bias. Meanwhile, retrieving training data is similar to utilizing prompts \cite{gupta-etal-2022-show}, but the worse performance compared to other knowledge formats suggests that selecting top-1 element is not optimal despite the relatively high recall. This is mostly due to that the retrieval results are noisy, as the small set of examples may contain slot types or values that are different from the ground truth. It is even less likely to find an example with exactly the same dialog state when the context is long. We leave further investigation by separating knowledge memory to support different knowledge sizes and external knowledge to future work.

\subsection{Analysis}
\label{analysis}
We study the relationship between retrieval and JGA in this section, and provide error analysis. 
We also analyze the detailed comparison between our method and D3ST.

\begin{figure}[h]
\centering
\includegraphics[width=\linewidth]{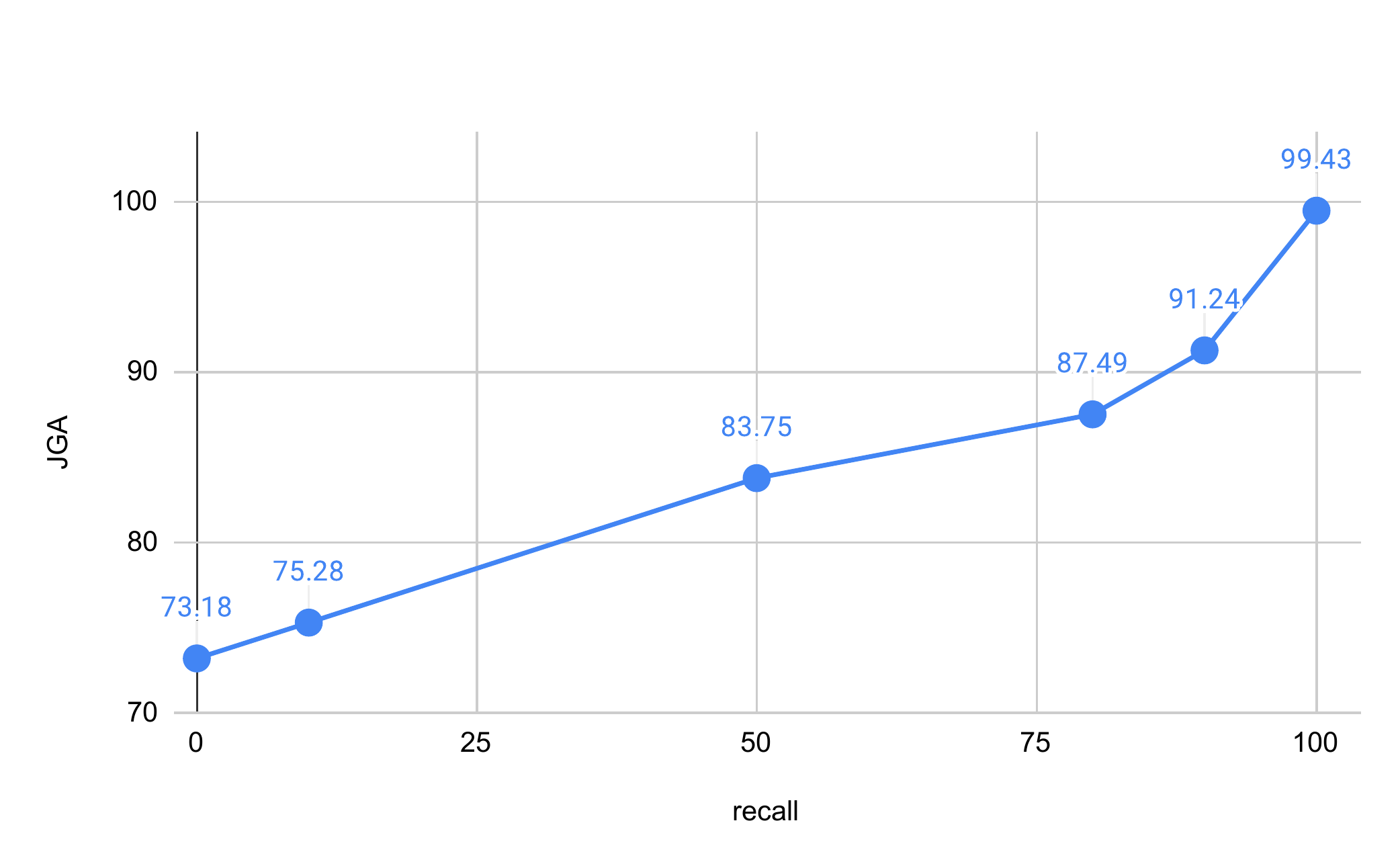}
\caption{JGA with controlled retrieval recall from sanity check experimented with T5-XXL on the full-data setting. Results show that similar to our findings, even noisy retrieval improves model performance on DST.}
\label{fig:jga_recall}
\end{figure}

\paragraph{Relationship between retrieval quality and JGA}
To understand the relationship between retrieval and the downstream task, we show JGA corresponding to recall in a controlled sanity check. Specifically, we randomly sample ground truth slot type-value pairs to match a target recall score and replace the rest dialog states with pairs uniformly sampled from the whole ontology (excluding the ground truth) without replacement as negative examples. Results (detailed in Figure \ref{fig:jga_recall}) show that with T5-XXL on the full-data setting, $50\%$ recall can significantly improve the model performance (83.75 JGA) while $90\%$ recall can result in 91.24 JGA. This suggests that a high recall for retrieval is critical to JGA, while the model remains robust against noisy retrieval results. It also indicates that a better retrieval method (such as an external one \citealt{lazaridou-etal-2022-internet}) may achieve better performance.
On the other hand, if we consider DST as a multi-class classification task with a retrieval module only, the model has to pick relevant elements from top-k, which is non-trivial.

We also consider separating retrieval from DST, i.e., train the model for retrieval first and then on DST. Results show that although the model can achieve 
$97.38\%$ recall, JGA actually drops to $70.33$ on the full-data setting with XXL. We conjecture the main reason to be that different from freezing retrieval index in previous question-answering work, knowledge such as ontology or training data are more homogeneous and thus being more sensitive.
This result is similar to our findings when training the two tasks jointly: retrieval metrics keep improving while JGA may drop with higher retrieval, even if we decrease the retrieval loss weight.

When we optimize separate parameters (i.e., two additional layers) for retrieval instead of the whole model, we observe slightly lower performance on JGA ($54.76$ compared to $55.32$ on $1\%$ data) and lower retrieval recall ($36$ compared to close to $46$).
Lastly, compared to top-30 with a JGA of $75.47$, we observe an absolute drop of $0.40$ for top-20, and $3.25$ for top-10. This indicates that compared to noise in precision, retrieving ground-truth elements for recall is more critical to JGA.

\paragraph{Comparison to D3ST}
D3ST decodes the sequence of dialog state based on the order of slot types provided in the prompt by data pre-processing. In comparison, the order of retrieved elements varies while the order of dialog state depends on the ground-truth annotation. In other words, similar to the seq2seq baseline, our method requires learning the annotation order for DST prediction. This makes it more challenging to train, especially when there are similar knowledge elements retrieved. This can be justified by the slightly lower JGA with the full data setting.  On the other hand, D3ST can be considered as a special setting of our grounding method where all knowledge elements are provided, and the DST generation model needs to implicitly detect relevant information and decode accordingly. We conjecture that the better performance on the few-shot setting over D3ST is due to that retrieving target elements while filtering noisy ones is easier than selecting corresponding knowledge, which can be shown from the high recall scores compared to lower JGA for D3ST. One future direction is to combine the benefits of the two worlds by utilizing the retrieved knowledge without length restriction.

\paragraph{Error analysis}
We found qualitatively that instead of ignoring retrieved elements as shown in previous research, the model does attend to retrieved slot-value pairs when decoding dialog states. The main errors are from noisy retrieval, where a very similar elements with a higher rank (thus closer to the context in $\mathbf{x'}$) than the ground truth knowledge may either stop the model from generating more states (i.e., missing target dialog states) or signal the model to generate the wrong elements directly. On the other hand, the model always predicts correctly if the ground truth are the most confident retrieved elements. To deal with the influence of attending only at the nearest few elements (which have the highest retrieval scores), we also experimented with randomly shuffling the retrieved knowledge but this results in lower scores ($71.0$ compared to $75.5$) because the model needs to denoise from potential top-k elements without any additional information.

\section{Conclusion}
In this paper, we propose to disentangle domain knowledge and encode knowledge as a prior to dialog state tracking. Compared to previous research of grounding on knowledge for factual generation, our method can be applied to multiple sources of knowledge in the task-oriented dialog understanding setting. We conduct experiments on the MultiWOZ dataset and show superior performance especially in the few-shot learning setting. We plan to apply our method on more general natural language understanding tasks in the future.

\section{Limitations}
In the experiments, we show model improvements over strong baselines. Despite the simplicity of the method, we acknowledge that the domain ontology is not always available since 
knowledge (e.g., non-categorical slots) may not be a closed set,
such as type+value in DST. However, this limitation can be lifted in two ways. Firstly, as shown in our experiments, retrieving slot type alone can also improve the model performance, which indicates that we may choose a knowledge base mixing type and type+value when the assumption that all values are predefined does not hold. Moreover, in most DST applications, the schema is specified before data collection and model training, where all target types and values need to match a database for information lookup. If the schema is unavailable, we may consider schema induction \cite{hudecek-etal-2021-discovering, yu-etal-2022-unsupervised} where we can build the schema before DST. We plan to investigate these directions in our future work.  

\section*{Acknowledgements}
We thank Abhinav Rastogi from Google Research, and anonymous reviewers for their constructive suggestions.

\bibliography{anthology,custom}

\begin{thebibliography}{45}
\expandafter\ifx\csname natexlab\endcsname\relax\def\natexlab#1{#1}\fi

\bibitem[{Borgeaud et~al.(2021)Borgeaud, Mensch, Hoffmann, Cai, Rutherford,
  Millican, van~den Driessche, Lespiau, Damoc, Clark, de~Las~Casas, Guy,
  Menick, Ring, Hennigan, Huang, Maggiore, Jones, Cassirer, Brock, Paganini,
  Irving, Vinyals, Osindero, Simonyan, Rae, Elsen, and
  Sifre}]{boregeaud-etal-2021-improving}
Sebastian Borgeaud, Arthur Mensch, Jordan Hoffmann, Trevor Cai, Eliza
  Rutherford, Katie Millican, George van~den Driessche, Jean{-}Baptiste
  Lespiau, Bogdan Damoc, Aidan Clark, Diego de~Las~Casas, Aurelia Guy, Jacob
  Menick, Roman Ring, Tom Hennigan, Saffron Huang, Loren Maggiore, Chris Jones,
  Albin Cassirer, Andy Brock, Michela Paganini, Geoffrey Irving, Oriol Vinyals,
  Simon Osindero, Karen Simonyan, Jack~W. Rae, Erich Elsen, and Laurent Sifre.
  2021.
\newblock \href {http://arxiv.org/abs/2112.04426} {Improving language models by
  retrieving from trillions of tokens}.
\newblock \emph{CoRR}, abs/2112.04426.

\bibitem[{Budzianowski et~al.(2018)Budzianowski, Wen, Tseng, Casanueva, Ultes,
  Ramadan, and Ga{\v{s}}i{\'c}}]{budzianowski-etal-2018-multiwoz}
Pawe{\l} Budzianowski, Tsung-Hsien Wen, Bo-Hsiang Tseng, I{\~n}igo Casanueva,
  Stefan Ultes, Osman Ramadan, and Milica Ga{\v{s}}i{\'c}. 2018.
\newblock \href {https://doi.org/10.18653/v1/D18-1547} {{M}ulti{WOZ} - a
  large-scale multi-domain {W}izard-of-{O}z dataset for task-oriented dialogue
  modelling}.
\newblock In \emph{Proceedings of the 2018 Conference on Empirical Methods in
  Natural Language Processing}, pages 5016--5026, Brussels, Belgium.
  Association for Computational Linguistics.

\bibitem[{Chen et~al.(2022)Chen, Liu, Moon, Sankar, Crook, and
  Wang}]{chen-etal-2022-ketod}
Zhiyu Chen, Bing Liu, Seungwhan Moon, Chinnadhurai Sankar, Paul Crook, and
  William~Yang Wang. 2022.
\newblock \href {https://doi.org/10.18653/v1/2022.findings-naacl.197} {{KETOD}:
  Knowledge-enriched task-oriented dialogue}.
\newblock In \emph{Findings of the Association for Computational Linguistics:
  NAACL 2022}, pages 2581--2593, Seattle, United States. Association for
  Computational Linguistics.

\bibitem[{Dinan et~al.(2019)Dinan, Roller, Shuster, Fan, Auli, and
  Weston}]{dinan-etal-2019-wizard}
Emily Dinan, Stephen Roller, Kurt Shuster, Angela Fan, Michael Auli, and Jason
  Weston. 2019.
\newblock {W}izard of {W}ikipedia: Knowledge-powered conversational agents.
\newblock In \emph{Proceedings of the International Conference on Learning
  Representations (ICLR)}.

\bibitem[{Du et~al.(2021)Du, He, Li, Yu, Pasupat, and Zhang}]{du-etal-2021-qa}
Xinya Du, Luheng He, Qi~Li, Dian Yu, Panupong Pasupat, and Yuan Zhang. 2021.
\newblock \href {https://doi.org/10.18653/v1/2021.acl-short.83} {{QA}-driven
  zero-shot slot filling with weak supervision pretraining}.
\newblock In \emph{Proceedings of the 59th Annual Meeting of the Association
  for Computational Linguistics and the 11th International Joint Conference on
  Natural Language Processing (Volume 2: Short Papers)}, pages 654--664,
  Online. Association for Computational Linguistics.

\bibitem[{Fan et~al.(2021)Fan, Gardent, Braud, and
  Bordes}]{fan-etal-2021-augmenting}
Angela Fan, Claire Gardent, Chlo{\'e} Braud, and Antoine Bordes. 2021.
\newblock \href {https://doi.org/10.1162/tacl_a_00356} {Augmenting transformers
  with {KNN}-based composite memory for dialog}.
\newblock \emph{Transactions of the Association for Computational Linguistics},
  9:82--99.

\bibitem[{Gonzalez et~al.(2019)Gonzalez, Augenstein, and
  Søgaard}]{gonzalez-etal-2019-retrievalbased}
Ana~Valeria Gonzalez, Isabelle Augenstein, and Anders Søgaard. 2019.
\newblock \href {http://arxiv.org/abs/1909.13717} {Retrieval-based
  goal-oriented dialogue generation}.

\bibitem[{Gupta et~al.(2022)Gupta, Lee, Zhao, Cao, Rastogi, and
  Wu}]{gupta-etal-2022-show}
Raghav Gupta, Harrison Lee, Jeffrey Zhao, Yuan Cao, Abhinav Rastogi, and
  Yonghui Wu. 2022.
\newblock \href {https://doi.org/10.18653/v1/2022.naacl-main.336} {Show,
  don{'}t tell: Demonstrations outperform descriptions for schema-guided
  task-oriented dialogue}.
\newblock In \emph{Proceedings of the 2022 Conference of the North American
  Chapter of the Association for Computational Linguistics: Human Language
  Technologies}, pages 4541--4549, Seattle, United States. Association for
  Computational Linguistics.

\bibitem[{Guu et~al.(2020)Guu, Lee, Tung, Pasupat, and
  Chang}]{guu-etal-2020-retrieval}
Kelvin Guu, Kenton Lee, Zora Tung, Panupong Pasupat, and Ming{-}Wei Chang.
  2020.
\newblock \href {http://proceedings.mlr.press/v119/guu20a.html} {Retrieval
  augmented language model pre-training}.
\newblock In \emph{Proceedings of the 37th International Conference on Machine
  Learning, {ICML} 2020, 13-18 July 2020, Virtual Event}, volume 119 of
  \emph{Proceedings of Machine Learning Research}, pages 3929--3938. {PMLR}.

\bibitem[{Hosseini-Asl et~al.(2020)Hosseini-Asl, McCann, Wu, Yavuz, and
  Socher}]{simpletod}
Ehsan Hosseini-Asl, Bryan McCann, Chien-Sheng Wu, Semih Yavuz, and Richard
  Socher. 2020.
\newblock \href
  {https://proceedings.neurips.cc/paper/2020/file/e946209592563be0f01c844ab2170f0c-Paper.pdf}
  {A simple language model for task-oriented dialogue}.
\newblock In \emph{Advances in Neural Information Processing Systems},
  volume~33, pages 20179--20191.

\bibitem[{Hude{\v{c}}ek et~al.(2021)Hude{\v{c}}ek, Du{\v{s}}ek, and
  Yu}]{hudecek-etal-2021-discovering}
Vojt{\v{e}}ch Hude{\v{c}}ek, Ond{\v{r}}ej Du{\v{s}}ek, and Zhou Yu. 2021.
\newblock \href {https://doi.org/10.18653/v1/2021.acl-long.189} {Discovering
  dialogue slots with weak supervision}.
\newblock In \emph{Proceedings of the 59th Annual Meeting of the Association
  for Computational Linguistics and the 11th International Joint Conference on
  Natural Language Processing (Volume 1: Long Papers)}, pages 2430--2442,
  Online. Association for Computational Linguistics.

\bibitem[{Izacard and Grave(2021)}]{izacard-grave-2021-distilling}
Gautier Izacard and Edouard Grave. 2021.
\newblock \href {https://openreview.net/forum?id=NTEz-6wysdb} {Distilling
  knowledge from reader to retriever for question answering}.
\newblock In \emph{International Conference on Learning Representations}.

\bibitem[{Jouppi et~al.(2017)Jouppi, Young, Patil, Patterson, Agrawal, Bajwa,
  Bates, Bhatia, Boden, Borchers, Boyle, Cantin, Chao, Clark, Coriell, Daley,
  Dau, Dean, Gelb, Ghaemmaghami, Gottipati, Gulland, Hagmann, Ho, Hogberg, Hu,
  Hundt, Hurt, Ibarz, Jaffey, Jaworski, Kaplan, Khaitan, Killebrew, Koch,
  Kumar, Lacy, Laudon, Law, Le, Leary, Liu, Lucke, Lundin, MacKean, Maggiore,
  Mahony, Miller, Nagarajan, Narayanaswami, Ni, Nix, Norrie, Omernick,
  Penukonda, Phelps, Ross, Ross, Salek, Samadiani, Severn, Sizikov, Snelham,
  Souter, Steinberg, Swing, Tan, Thorson, Tian, Toma, Tuttle, Vasudevan,
  Walter, Wang, Wilcox, and Yoon}]{jouppi-etal-2017-in}
Norman~P. Jouppi, Cliff Young, Nishant Patil, David Patterson, Gaurav Agrawal,
  Raminder Bajwa, Sarah Bates, Suresh Bhatia, Nan Boden, Al~Borchers, Rick
  Boyle, Pierre-luc Cantin, Clifford Chao, Chris Clark, Jeremy Coriell, Mike
  Daley, Matt Dau, Jeffrey Dean, Ben Gelb, Tara~Vazir Ghaemmaghami, Rajendra
  Gottipati, William Gulland, Robert Hagmann, C.~Richard Ho, Doug Hogberg, John
  Hu, Robert Hundt, Dan Hurt, Julian Ibarz, Aaron Jaffey, Alek Jaworski,
  Alexander Kaplan, Harshit Khaitan, Daniel Killebrew, Andy Koch, Naveen Kumar,
  Steve Lacy, James Laudon, James Law, Diemthu Le, Chris Leary, Zhuyuan Liu,
  Kyle Lucke, Alan Lundin, Gordon MacKean, Adriana Maggiore, Maire Mahony,
  Kieran Miller, Rahul Nagarajan, Ravi Narayanaswami, Ray Ni, Kathy Nix, Thomas
  Norrie, Mark Omernick, Narayana Penukonda, Andy Phelps, Jonathan Ross, Matt
  Ross, Amir Salek, Emad Samadiani, Chris Severn, Gregory Sizikov, Matthew
  Snelham, Jed Souter, Dan Steinberg, Andy Swing, Mercedes Tan, Gregory
  Thorson, Bo~Tian, Horia Toma, Erick Tuttle, Vijay Vasudevan, Richard Walter,
  Walter Wang, Eric Wilcox, and Doe~Hyun Yoon. 2017.
\newblock \href {https://doi.org/10.1145/3079856.3080246} {In-datacenter
  performance analysis of a tensor processing unit}.
\newblock In \emph{Proceedings of the 44th Annual International Symposium on
  Computer Architecture}, ISCA '17, page 1–12, New York, NY, USA. Association
  for Computing Machinery.

\bibitem[{Karpukhin et~al.(2020)Karpukhin, Oguz, Min, Lewis, Wu, Edunov, Chen,
  and Yih}]{karpukhin-etal-2020-dense}
Vladimir Karpukhin, Barlas Oguz, Sewon Min, Patrick Lewis, Ledell Wu, Sergey
  Edunov, Danqi Chen, and Wen-tau Yih. 2020.
\newblock \href {https://doi.org/10.18653/v1/2020.emnlp-main.550} {Dense
  passage retrieval for open-domain question answering}.
\newblock In \emph{Proceedings of the 2020 Conference on Empirical Methods in
  Natural Language Processing (EMNLP)}, pages 6769--6781, Online. Association
  for Computational Linguistics.

\bibitem[{Khandelwal et~al.(2021)Khandelwal, Fan, Jurafsky, Zettlemoyer, and
  Lewis}]{khandelwal-etal-2021-nearest}
Urvashi Khandelwal, Angela Fan, Dan Jurafsky, Luke Zettlemoyer, and Mike Lewis.
  2021.
\newblock \href {https://openreview.net/forum?id=7wCBOfJ8hJM} {Nearest neighbor
  machine translation}.
\newblock In \emph{International Conference on Learning Representations}.

\bibitem[{Khattab and Zaharia(2020)}]{khattab-etal-2020-colbert}
Omar Khattab and Matei Zaharia. 2020.
\newblock \href {https://doi.org/10.1145/3397271.3401075} {\emph{ColBERT:
  Efficient and Effective Passage Search via Contextualized Late Interaction
  over BERT}}, page 39–48. Association for Computing Machinery, New York, NY,
  USA.

\bibitem[{Kim et~al.(2020)Kim, Eric, Gopalakrishnan, Hedayatnia, Liu, and
  Hakkani-Tur}]{kim-etal-2020-beyond}
Seokhwan Kim, Mihail Eric, Karthik Gopalakrishnan, Behnam Hedayatnia, Yang Liu,
  and Dilek Hakkani-Tur. 2020.
\newblock \href {https://aclanthology.org/2020.sigdial-1.35} {Beyond domain
  {API}s: Task-oriented conversational modeling with unstructured knowledge
  access}.
\newblock In \emph{Proceedings of the 21th Annual Meeting of the Special
  Interest Group on Discourse and Dialogue}, pages 278--289, 1st virtual
  meeting. Association for Computational Linguistics.

\bibitem[{Komeili et~al.(2022)Komeili, Shuster, and
  Weston}]{komeili-etal-2022-internet}
Mojtaba Komeili, Kurt Shuster, and Jason Weston. 2022.
\newblock \href {https://doi.org/10.18653/v1/2022.acl-long.579}
  {{I}nternet-augmented dialogue generation}.
\newblock In \emph{Proceedings of the 60th Annual Meeting of the Association
  for Computational Linguistics (Volume 1: Long Papers)}, pages 8460--8478,
  Dublin, Ireland. Association for Computational Linguistics.

\bibitem[{Lazaridou et~al.(2022)Lazaridou, Gribovskaya, Stokowiec, and
  Grigorev}]{lazaridou-etal-2022-internet}
Angeliki Lazaridou, Elena Gribovskaya, Wojciech Stokowiec, and Nikolai
  Grigorev. 2022.
\newblock \href {https://doi.org/10.48550/ARXIV.2203.05115} {Internet-augmented
  language models through few-shot prompting for open-domain question
  answering}.

\bibitem[{Lee et~al.(2021)Lee, Cheng, and Ostendorf}]{lee-etal-2021-dialogue}
Chia-Hsuan Lee, Hao Cheng, and Mari Ostendorf. 2021.
\newblock \href {https://doi.org/10.18653/v1/2021.emnlp-main.404} {Dialogue
  state tracking with a language model using schema-driven prompting}.
\newblock In \emph{Proceedings of the 2021 Conference on Empirical Methods in
  Natural Language Processing}, pages 4937--4949, Online and Punta Cana,
  Dominican Republic. Association for Computational Linguistics.

\bibitem[{Lee et~al.(2019)Lee, Chang, and Toutanova}]{lee-etal-2019-latent}
Kenton Lee, Ming-Wei Chang, and Kristina Toutanova. 2019.
\newblock \href {https://doi.org/10.18653/v1/P19-1612} {Latent retrieval for
  weakly supervised open domain question answering}.
\newblock In \emph{Proceedings of the 57th Annual Meeting of the Association
  for Computational Linguistics}, pages 6086--6096, Florence, Italy.
  Association for Computational Linguistics.

\bibitem[{Lewis et~al.(2020)Lewis, Perez, Piktus, Petroni, Karpukhin, Goyal,
  K\"{u}ttler, Lewis, Yih, Rockt\"{a}schel, Riedel, and
  Kiela}]{lewis-etal-2020-retrieval}
Patrick Lewis, Ethan Perez, Aleksandra Piktus, Fabio Petroni, Vladimir
  Karpukhin, Naman Goyal, Heinrich K\"{u}ttler, Mike Lewis, Wen-tau Yih, Tim
  Rockt\"{a}schel, Sebastian Riedel, and Douwe Kiela. 2020.
\newblock \href
  {https://proceedings.neurips.cc/paper/2020/file/6b493230205f780e1bc26945df7481e5-Paper.pdf}
  {Retrieval-augmented generation for knowledge-intensive nlp tasks}.
\newblock In \emph{Advances in Neural Information Processing Systems},
  volume~33, pages 9459--9474.

\bibitem[{Madotto et~al.(2020)Madotto, Cahyawijaya, Winata, Xu, Liu, Lin, and
  Fung}]{madotto-etal-2020-learning}
Andrea Madotto, Samuel Cahyawijaya, Genta~Indra Winata, Yan Xu, Zihan Liu,
  Zhaojiang Lin, and Pascale Fung. 2020.
\newblock \href {https://doi.org/10.18653/v1/2020.findings-emnlp.215} {Learning
  knowledge bases with parameters for task-oriented dialogue systems}.
\newblock In \emph{Findings of the Association for Computational Linguistics:
  EMNLP 2020}, pages 2372--2394, Online. Association for Computational
  Linguistics.

\bibitem[{Ni et~al.(2021)Ni, Qu, Lu, Dai, {\'{A}}brego, Ma, Zhao, Luan, Hall,
  Chang, and Yang}]{ni-etal-2021-large}
Jianmo Ni, Chen Qu, Jing Lu, Zhuyun Dai, Gustavo~Hern{\'{a}}ndez {\'{A}}brego,
  Ji~Ma, Vincent~Y. Zhao, Yi~Luan, Keith~B. Hall, Ming{-}Wei Chang, and Yinfei
  Yang. 2021.
\newblock \href {http://arxiv.org/abs/2112.07899} {Large dual encoders are
  generalizable retrievers}.
\newblock \emph{CoRR}, abs/2112.07899.

\bibitem[{Pasupat et~al.(2021)Pasupat, Zhang, and
  Guu}]{pasupat-etal-2021-controllable}
Panupong Pasupat, Yuan Zhang, and Kelvin Guu. 2021.
\newblock \href {https://doi.org/10.18653/v1/2021.emnlp-main.607} {Controllable
  semantic parsing via retrieval augmentation}.
\newblock In \emph{Proceedings of the 2021 Conference on Empirical Methods in
  Natural Language Processing}, pages 7683--7698, Online and Punta Cana,
  Dominican Republic. Association for Computational Linguistics.

\bibitem[{Peng et~al.(2020)Peng, Li, Li, Shayandeh, Liden, and
  Gao}]{peng-etal-2020-soloist}
Baolin Peng, Chunyuan Li, Jinchao Li, Shahin Shayandeh, Lars Liden, and
  Jianfeng Gao. 2020.
\newblock Soloist: Building task bots at scale with transfer learning and
  machine teaching.
\newblock \emph{arXiv preprint arXiv:2005.05298}.

\bibitem[{Radford et~al.(2019)Radford, Wu, Child, Luan, Amodei, and
  Sutskever}]{gpt2}
Alec Radford, Jeff Wu, Rewon Child, David Luan, Dario Amodei, and Ilya
  Sutskever. 2019.
\newblock Language models are unsupervised multitask learners.
\newblock \emph{CoRR}.

\bibitem[{Raffel et~al.(2020)Raffel, Shazeer, Roberts, Lee, Narang, Matena,
  Zhou, Li, and Liu}]{t5}
Colin Raffel, Noam Shazeer, Adam Roberts, Katherine Lee, Sharan Narang, Michael
  Matena, Yanqi Zhou, Wei Li, and Peter~J. Liu. 2020.
\newblock \href {http://jmlr.org/papers/v21/20-074.html} {Exploring the limits
  of transfer learning with a unified text-to-text transformer}.
\newblock \emph{Journal of Machine Learning Research}, 21(140):1--67.

\bibitem[{Rastogi et~al.(2020)Rastogi, Zang, Sunkara, Gupta, and
  Khaitan}]{rastogi-etal-2020-towards}
Abhinav Rastogi, Xiaoxue Zang, Srinivas Sunkara, Raghav Gupta, and Pranav
  Khaitan. 2020.
\newblock \href {https://doi.org/10.1609/aaai.v34i05.6394} {Towards scalable
  multi-domain conversational agents: The schema-guided dialogue dataset}.
\newblock \emph{Proceedings of the AAAI Conference on Artificial Intelligence},
  34(05):8689--8696.

\bibitem[{Ren et~al.(2018)Ren, Xie, Chen, and Yu}]{ren-etal-2018-towards}
Liliang Ren, Kaige Xie, Lu~Chen, and Kai Yu. 2018.
\newblock \href {https://doi.org/10.18653/v1/D18-1299} {Towards universal
  dialogue state tracking}.
\newblock In \emph{Proceedings of the 2018 Conference on Empirical Methods in
  Natural Language Processing}, pages 2780--2786, Brussels, Belgium.
  Association for Computational Linguistics.

\bibitem[{Robertson and Zaragoza(2009)}]{robertson-zaragoza-2009-bm25}
Stephen Robertson and Hugo Zaragoza. 2009.
\newblock \href {https://doi.org/10.1561/1500000019} {The probabilistic
  relevance framework: Bm25 and beyond}.
\newblock \emph{Found. Trends Inf. Retr.}, 3(4):333–389.

\bibitem[{Shuster et~al.(2022)Shuster, Komeili, Adolphs, Roller, Szlam, and
  Weston}]{shuster-etal-2022-language}
Kurt Shuster, Mojtaba Komeili, Leonard Adolphs, Stephen Roller, Arthur Szlam,
  and Jason Weston. 2022.
\newblock \href {https://doi.org/10.48550/ARXIV.2203.13224} {Language models
  that seek for knowledge: Modular search \& generation for dialogue and prompt
  completion}.

\bibitem[{Thoppilan et~al.(2022)Thoppilan, Freitas, Hall, Shazeer,
  Kulshreshtha, Cheng, Jin, Bos, Baker, Du, Li, Lee, Zheng, Ghafouri, Menegali,
  Huang, Krikun, Lepikhin, Qin, Chen, Xu, Chen, Roberts, Bosma, Zhou, Chang,
  Krivokon, Rusch, Pickett, Meier{-}Hellstern, Morris, Doshi, Santos, Duke,
  Soraker, Zevenbergen, Prabhakaran, Diaz, Hutchinson, Olson, Molina,
  Hoffman{-}John, Lee, Aroyo, Rajakumar, Butryna, Lamm, Kuzmina, Fenton, Cohen,
  Bernstein, Kurzweil, Aguera{-}Arcas, Cui, Croak, Chi, and Le}]{lamda}
Romal Thoppilan, Daniel~De Freitas, Jamie Hall, Noam Shazeer, Apoorv
  Kulshreshtha, Heng{-}Tze Cheng, Alicia Jin, Taylor Bos, Leslie Baker, Yu~Du,
  YaGuang Li, Hongrae Lee, Huaixiu~Steven Zheng, Amin Ghafouri, Marcelo
  Menegali, Yanping Huang, Maxim Krikun, Dmitry Lepikhin, James Qin, Dehao
  Chen, Yuanzhong Xu, Zhifeng Chen, Adam Roberts, Maarten Bosma, Yanqi Zhou,
  Chung{-}Ching Chang, Igor Krivokon, Will Rusch, Marc Pickett, Kathleen~S.
  Meier{-}Hellstern, Meredith~Ringel Morris, Tulsee Doshi, Renelito~Delos
  Santos, Toju Duke, Johnny Soraker, Ben Zevenbergen, Vinodkumar Prabhakaran,
  Mark Diaz, Ben Hutchinson, Kristen Olson, Alejandra Molina, Erin
  Hoffman{-}John, Josh Lee, Lora Aroyo, Ravi Rajakumar, Alena Butryna, Matthew
  Lamm, Viktoriya Kuzmina, Joe Fenton, Aaron Cohen, Rachel Bernstein, Ray
  Kurzweil, Blaise Aguera{-}Arcas, Claire Cui, Marian Croak, Ed~H. Chi, and
  Quoc Le. 2022.
\newblock \href {http://arxiv.org/abs/2201.08239} {Lamda: Language models for
  dialog applications}.
\newblock \emph{CoRR}, abs/2201.08239.

\bibitem[{Thulke et~al.(2021)Thulke, Daheim, Dugast, and
  Ney}]{thulke-etal-2021-efficient}
David Thulke, Nico Daheim, Christian Dugast, and Hermann Ney. 2021.
\newblock \href {http://arxiv.org/abs/2102.04643} {Efficient retrieval
  augmented generation from unstructured knowledge for task-oriented dialog}.

\bibitem[{Weston et~al.(2018)Weston, Dinan, and
  Miller}]{weston-etal-2018-retrieve}
Jason Weston, Emily Dinan, and Alexander Miller. 2018.
\newblock \href {https://doi.org/10.18653/v1/W18-5713} {Retrieve and refine:
  Improved sequence generation models for dialogue}.
\newblock In \emph{Proceedings of the 2018 {EMNLP} Workshop {SCAI}: The 2nd
  International Workshop on Search-Oriented Conversational {AI}}, pages 87--92,
  Brussels, Belgium. Association for Computational Linguistics.

\bibitem[{Wu et~al.(2020)Wu, Hoi, Socher, and Xiong}]{wu-etal-2020-tod}
Chien-Sheng Wu, Steven~C.H. Hoi, Richard Socher, and Caiming Xiong. 2020.
\newblock \href {https://doi.org/10.18653/v1/2020.emnlp-main.66} {{TOD}-{BERT}:
  Pre-trained natural language understanding for task-oriented dialogue}.
\newblock In \emph{Proceedings of the 2020 Conference on Empirical Methods in
  Natural Language Processing (EMNLP)}, pages 917--929, Online. Association for
  Computational Linguistics.

\bibitem[{Wu et~al.(2019)Wu, Madotto, Hosseini-Asl, Xiong, Socher, and
  Fung}]{wu-etal-2019-transferable}
Chien-Sheng Wu, Andrea Madotto, Ehsan Hosseini-Asl, Caiming Xiong, Richard
  Socher, and Pascale Fung. 2019.
\newblock \href {https://doi.org/10.18653/v1/P19-1078} {Transferable
  multi-domain state generator for task-oriented dialogue systems}.
\newblock In \emph{Proceedings of the 57th Annual Meeting of the Association
  for Computational Linguistics}, pages 808--819, Florence, Italy. Association
  for Computational Linguistics.

\bibitem[{Yang et~al.(2021)Yang, Li, and Quan}]{yang-etal-2021-ubar}
Yunyi Yang, Yunhao Li, and Xiaojun Quan. 2021.
\newblock \href {https://ojs.aaai.org/index.php/AAAI/article/view/17674} {Ubar:
  Towards fully end-to-end task-oriented dialog system with gpt-2}.
\newblock \emph{Proceedings of the AAAI Conference on Artificial Intelligence},
  35(16):14230--14238.

\bibitem[{Yao et~al.(2021)Yao, Zheng, Yang, and Yang}]{yao-etal-2021-tlm}
Xingcheng Yao, Yanan Zheng, Xiaocong Yang, and Zhilin Yang. 2021.
\newblock Nlp from scratch without large-scale pretraining: A simple and
  efficient framework.
\newblock In \emph{ICML}.

\bibitem[{Ye et~al.(2021)Ye, Manotumruksa, and Yilmaz}]{ye-etal-2021-multiwoz}
Fanghua Ye, Jarana Manotumruksa, and Emine Yilmaz. 2021.
\newblock \href {http://arxiv.org/abs/2104.00773} {Multiwoz 2.4: {A}
  multi-domain task-oriented dialogue dataset with essential annotation
  corrections to improve state tracking evaluation}.
\newblock \emph{CoRR}, abs/2104.00773.

\bibitem[{Yu et~al.(2021)Yu, He, Zhang, Du, Pasupat, and
  Li}]{yu-etal-2021-shot}
Dian Yu, Luheng He, Yuan Zhang, Xinya Du, Panupong Pasupat, and Qi~Li. 2021.
\newblock \href {https://doi.org/10.18653/v1/2021.naacl-main.59} {Few-shot
  intent classification and slot filling with retrieved examples}.
\newblock In \emph{Proceedings of the 2021 Conference of the North American
  Chapter of the Association for Computational Linguistics: Human Language
  Technologies}, pages 734--749, Online. Association for Computational
  Linguistics.

\bibitem[{Yu et~al.(2022)Yu, Wang, Cao, Shafran, Shafey, and
  Soltau}]{yu-etal-2022-unsupervised}
Dian Yu, Mingqiu Wang, Yuan Cao, Izhak Shafran, Laurent Shafey, and Hagen
  Soltau. 2022.
\newblock \href {https://doi.org/10.18653/v1/2022.naacl-main.86} {Unsupervised
  slot schema induction for task-oriented dialog}.
\newblock In \emph{Proceedings of the 2022 Conference of the North American
  Chapter of the Association for Computational Linguistics: Human Language
  Technologies}, pages 1174--1193, Seattle, United States. Association for
  Computational Linguistics.

\bibitem[{Zhang et~al.(2021)Zhang, Sun, Gao, Fang, Brockett, Galley, Gao, and
  Dolan}]{zhang-etal-2021-jointRetrieval}
Yizhe Zhang, Siqi Sun, Xiang Gao, Yuwei Fang, Chris Brockett, Michel Galley,
  Jianfeng Gao, and Bill Dolan. 2021.
\newblock Joint retrieval and generation training for grounded text generation.
\newblock \emph{arXiv preprint arXiv:2105.06597}.

\bibitem[{Zhao et~al.(2022)Zhao, Gupta, Cao, Yu, Wang, Lee, Rastogi, Shafran,
  and Wu}]{zhao-etal-2022-description}
Jeffrey Zhao, Raghav Gupta, Yuan Cao, Dian Yu, Mingqiu Wang, Harrison Lee,
  Abhinav Rastogi, Izhak Shafran, and Yonghui Wu. 2022.
\newblock \href {http://arxiv.org/abs/2201.08904} {Description-driven
  task-oriented dialog modeling}.
\newblock \emph{CoRR}, abs/2201.08904.

\bibitem[{Zhou and Small(2019)}]{zhou-small-2019-multi}
Li~Zhou and Kevin Small. 2019.
\newblock \href {http://arxiv.org/abs/1911.06192} {Multi-domain dialogue state
  tracking as dynamic knowledge graph enhanced question answering}.
\newblock \emph{CoRR}, abs/1911.06192.

\end{thebibliography}
\bibliographystyle{acl_natbib}

\clearpage

\appendix

\end{document}